# Satellite-Net: Automatic Extraction of Land Cover Indicators from Satellite Imagery by Deep Learning

Eleonora Bernasconi

Sapienza, University of Rome, Italy

Francesco Pugliese

Italian National Institute of Statistics

Diego Zardetto

Italian National Institute of Statistics

Monica Scannapieco

Italian National Institute of Statistics

1. INTRODUCTION

In this paper we address the challenge of land cover classification for satellite images via Deep Learning (DL). Land Cover aims to detect the physical characteristics of the territory and estimate the percentage of land occupied by a certain category of entities: vegetation, residential buildings, industrial areas, forest areas, rivers, lakes, etc. DL is a new paradigm for Big Data analytics and in particular for Computer Vision. The application of DL in images classification for land cover purposes has a great potential owing to the high degree of automation and computing performance. In particular, the invention of Convolution Neural Networks (CNNs) was a fundament for the advancements in this field. In [1], the Satellite Task Team of the UN Global Working Group describes the results achieved so far with respect to the use of earth observation for Official Statistics. However, in that study, CNNs have not yet been explored for automatic classification of imagery. This work investigates the usage of CNNs for the estimation of land cover indicators, providing evidence of the first promising results. In particular, the paper proposes a customized model, called Satellite-Net, able to reach an accuracy level up to 98% on test sets.

2. METHODS

2.1. Training Dataset and Model Architecture

Deep Learning methods for land cover estimation require large labelled-data sets to train the underlying deep neural networks. For this challenging task, we use the openly and freely accessible Sentinel-2A satellite images set, provided within the scope of the Earth observation program Copernicus. Images were enclosed in a dataset called EuroSAT [2] and used to train the CNN for the satellite images recognition. The dataset consists of 27,000 images labelled with ten different classes. Each image is in RGB colours and has a 64x64-pixel size. The ten folders that divide the dataset are listed in Fig. 1.

| 0 | Annual Crop | 5 | Pasture |
|---|---|---|---|
| 1 | Forest | 6 | Permanent Crop |
| 2 | Herbaceous Vegetation | 7 | Residential |
| 3 | Highway | 8 | River |
| 4 | Industrial | 9 | Sea Lake |

Figure 1. Categories of the EuroSAT dataset

The original dataset has been split into the following three parts: training-set (80%), validation-set (10%) and test-set (10%), in order to perform the training, validation and test, respectively. We designed a new model called "Satellite-Net" whose topology is depicted in Fig. 2. This artificial neural network is basically composed of four Convolutional Layers, six Batch Normalization Layers and six Activation Layers with a ReLu (Rectified Linear Unit) activation function. Furthermore, we included three Max-Pooling Layers with 2-sized strides, three Dropout Layers with a 0.5 drop probability. Finally, we can see one Flatten Layer, two Dense Layers and a Softmax Layer [3]. To cope with the lack in size of the dataset and thus allow the network to achieve better accuracies, a Data Augmentation was implemented by exclusively regulating the following variables: zoom range within the interval [0.8, 1.2] and rotation range within [0, 30]. Finally we set horizontal flip = True and vertical flip = True.

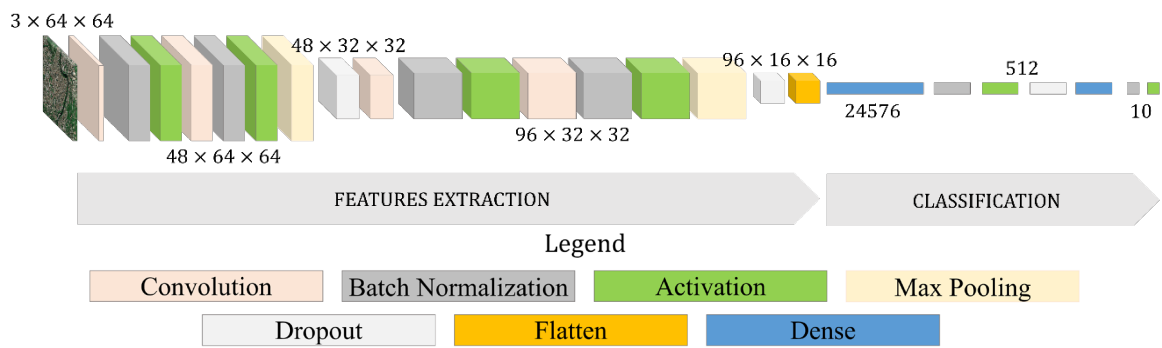

Figure 2. Satellite-Net model

The CNN has been trained for 300 epochs on an NVIDIA GTX-1080 achieving a time performance of 35s for each epoch. Hyper-parameters of the training process are: optimizer type: Adam, learning rate: 0.01 and batch-size: 32.

## 3. RESULTS

### 3.1. Model Performance

In Fig. 3a the Confusion Matrix of Satellite-Net classifier is reported which was generated during the test phase.

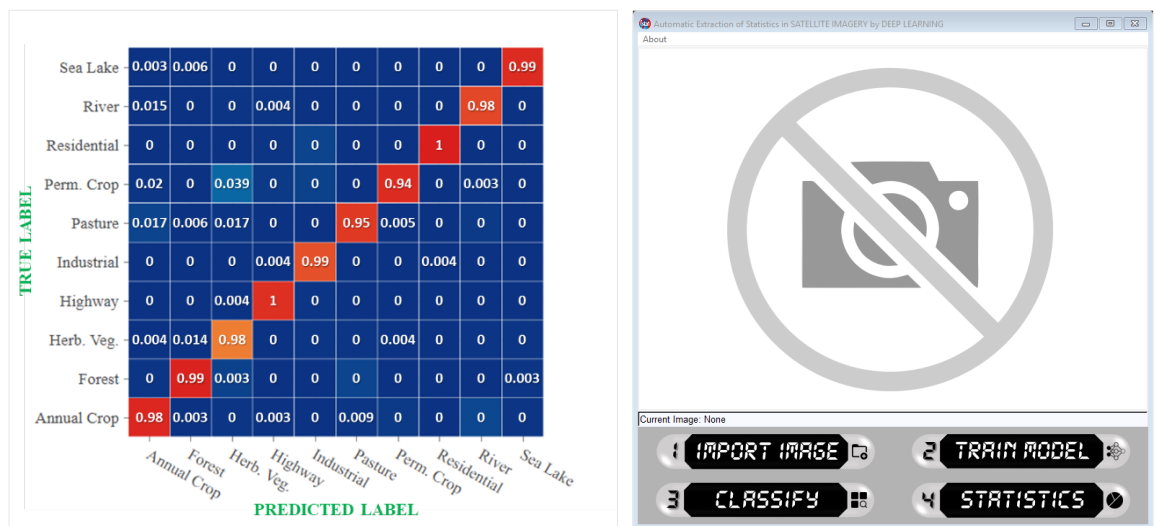

Figure 3. a) Confusion Matrix b) Interface

As we previously stated the accuracy achieved by our model on a test set is around 98.22%. The considered test set is made up of 2700 images (270 for each category) randomly extracted from the original EuroSAT dataset. These results are rather encouraging since we reached same accuracy of previous works [1] with less computational workload and a leaner neural model. Furthermore, this lead to a shortening in training and testing times.

A user-friendly interface (see Fig. 3b) has been conceived in Python-Tkinter library with four buttons, namely: "import" for reading the images; "training" for training Satellite-Net model; "classify" to predict label for each tile of import image; "statistics" for the extraction of land cover percentages.

### 3.2. Using the Model in a Production Setting

In a production setting, our trained CNN must prove able to predict land cover from satellite (or aerial) images that are much bigger (in both size and resolution) than those provided by the EuroSAT dataset. To this purpose, we implemented a Sliding-Window Scanning function. This function behaves as follows. First, it divides each high resolution satellite image into tiles of size (64 x 64). Then, it executes a scrolling procedure in order to visit and analyze each image tile once, with no overlaps (stride = 64). This procedure classifies one tile at a time and attaches the predicted land cover label (i.e. highway, forest, residential, etc.) to the corresponding area of the original image[1]. Therefore, the output of the whole process is a classification matrix: each element of this matrix corresponds to a tile of the original image and stores its predicted land cover class. Given this classification matrix, one can easily estimate the share of each land cover class for the whole territory depicted by the input satellite image (as well as for sub-regions of it, provided they can be obtained by aggregation of tiles). In fact, the share of any land cover class directly results from the relative frequency of the corresponding label within the matrix.

For the sake of clarity, let us consider a satellite image depicting the south-east side of Puglia territory in Italy (see Fig. 4a). This picture comes in a Jpeg2000 format from Sentinel-2A, with a size of (10,980 x 10,980) pixels.

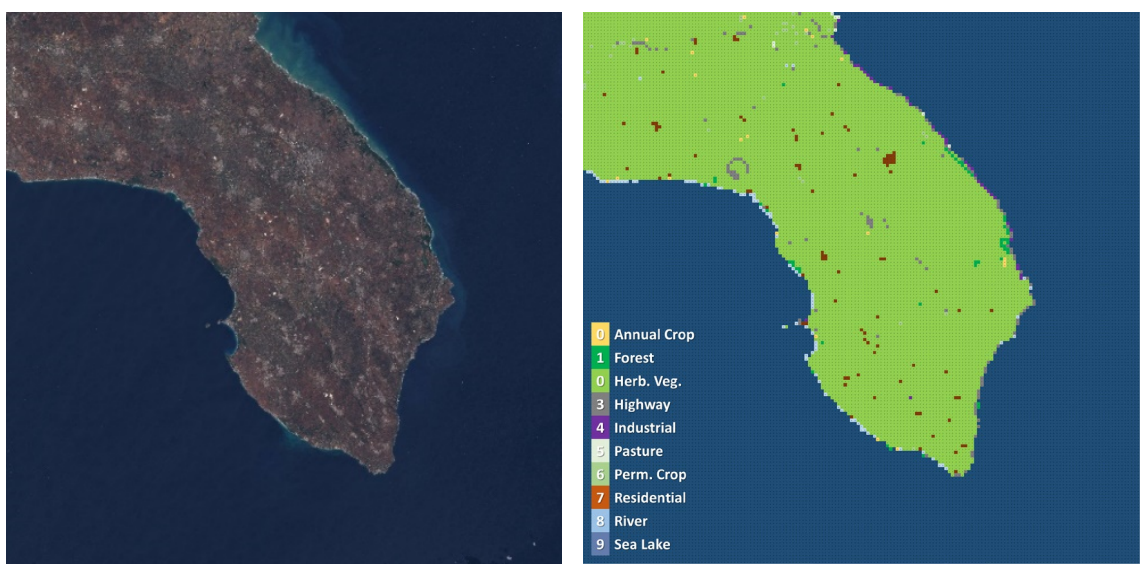

Figure 4. a) Satellite image of Puglia. b) Land cover of Puglia

---

[1] Of course, the cells of the matrix are arranged in such a way that the corresponding tiles preserve the spatial structure of the original satellite image.

According to the process described above, the image was partitioned into 171 x 171 tiles of (nearly) constant size (64 x 64), yielding a 171x171 classification matrix. This matrix contains all the predicted labels returned during the image scanning stage, when our trained CNN was asked to classify the tiles one at a time. As a consequence, the matrix can be used to depict a land cover map, where each land cover class corresponds to a different colour, as shown in Fig. 3b.

By computing the relative frequency of, say, the "Highway" label within the classification matrix, we obtain a raw estimate of the land cover share for highways (see Table 1). Note that this estimated land cover share is raw in that it is relative to the whole area depicted by the input image, including the sea. To derive the actual land cover share of highways, we would have to exclude from the classification matrix the cells that are linked to sea tiles. This task can be easily accomplished, either by suitably cropping the input image or by identifying sea tiles through a GIS. Table 1 below reports the raw land cover percentages referred to the picture in Fig. 4.

Table 1. Raw land cover estimate of Fig. 4

| Annual Crop | 0.07% |
|---|---|
| Forest | 0.17% |
| Herbaceous vegetation | 33.40% |
| Highway | 0.48% |
| Industrial | 0.13% |
| Pasture | 0.03% |
| Permanent Crop | 0.11% |
| Residential | 0.32% |
| River | 0.29% |
| Sea Lake | 65.01% |

## 4. CONCLUSIONS

Convolutional Neural Networks prove to be an effective and efficient way to approach remote sensing problems and in particular, the problem of labelling satellite images, because they are able to recognize and classify them.

When evaluated internally, that is on test sets derived from EuroSAT data, our CNN model shows a very high performance in terms of accuracy. Our next step will be to evaluate its performance externally, by testing its predictions on a set of labelled aerial images of the Italian territory which are typical of real Istat production scenarios.